%% file: journal.tex
\documentclass[journal]{journal}

\newcommand{\bfsection}[1]{\vspace*{0.1cm}\noindent\textbf{#1.}}
\newcommand*\rot{\rotatebox{90}}
\usepackage{graphicx}
\ifCLASSINFOpdf
\else
\fi
\usepackage[cmex10]{amsmath}
\begin{document}
%
\title{Nighttime Dehaze-Enhancement}
%
%
%

\author{
Harshan~Baskar\IEEEauthorrefmark{1},
Anirudh~S~Chakravarthy\IEEEauthorrefmark{1},
Prateek~Garg\IEEEauthorrefmark{1},
Divyam~Goel\IEEEauthorrefmark{1},
Abhijith~S~Raj\IEEEauthorrefmark{1},
Kshitij~Kumar\IEEEauthorrefmark{1}, \\
Lakshya\IEEEauthorrefmark{1},
Ravichandra~Parvatham\IEEEauthorrefmark{1},
V~Sushant\IEEEauthorrefmark{1},
Bijay~Kumar~Rout\IEEEauthorrefmark{1}\IEEEauthorrefmark{2}
\\
\IEEEauthorrefmark{1} Center for Robotics and Intelligent Systems, BITS Pilani, India \\
\IEEEauthorrefmark{2} Department of Mechanical Engineering, BITS Pilani, India


}

%
%

\markboth{Journal of \LaTeX\ Class Files,~Vol.~6, No.~1, January~2007}%
{Shell \MakeLowercase{\textit{et al.}}: Bare Demo of IEEEtran.cls for Journals}
%



\maketitle
\thispagestyle{empty}

\begin{abstract}
In this paper, we introduce a new computer vision task called nighttime dehaze-enhancement. This task aims to jointly perform dehazing and lightness enhancement. Our task fundamentally differs from nighttime dehazing -- our goal is to \textit{jointly} dehaze and enhance scenes, while nighttime dehazing aims to dehaze scenes \textit{under a nighttime setting}.
    In order to facilitate further research on this task, we release a new benchmark dataset called Reside-$\beta$ Night dataset, consisting of 4122 nighttime hazed images from 2061 scenes and 2061 ground truth images. Moreover, we also propose a new network called NDENet (Nighttime Dehaze-Enhancement Network), which jointly performs dehazing and low-light enhancement in an end-to-end manner. We evaluate our method on the proposed benchmark and achieve SSIM of 0.8962 and PSNR of 26.25. We also compare our network with other baseline networks on our benchmark to demonstrate the effectiveness of our approach. We believe that nighttime dehaze-enhancement is an essential task particularly for autonomous navigation applications, and hope that our work will open up new frontiers in research. Our dataset and code will be made publicly available upon acceptance of our paper.
\end{abstract}


%
\IEEEpeerreviewmaketitle

\section{Introduction}
%
%
%
%
\IEEEPARstart{T}{he} presence of dust, smoke, mist, and other particles in the atmosphere often leads to blurring, colour distortion, and low contrast in images. Images affected by haze are more challenging for downstream tasks. Image dehazing is crucial in applications where real-time image capture is necessary, such as autonomous navigation, remote sensing, surveillance, etc.

Daytime dehazing is a well-explored task in computer vision. Traditional approaches for daytime dehazing require image priors in order to perform dehazing. However, these priors don't apply to images captured in the nighttime. For example, the dark channel prior~\cite{he2010dcp} doesn't hold when the colours of objects in scene are very similar to the atmospheric lighting, as is the case in a nighttime scene. Therefore, learning-based dehazing approaches were introduced to overcome these issues.

\input{fig-tex/teaser}

However, dehazing is much more challenging when there is insufficient lighting during image capture. Such a low-light setting leads to low contrast, loss of essential details and thus degrades visibility and image quality. In such settings, even learning-based daytime dehazing algorithms fail~\cite{tang2021nighttime}. When nighttime images are used for downstream computer vision tasks in applications such as autonomous navigation, the quality must be restored. The loss of details due to low contrast, insufficient lighting, and haze pose serious challenges for such practical applications. Hence, nighttime dehazing has drawn the attention of many researchers and several approaches have been proposed to solve this task. However, nighttime dehazing approaches still leave object details unclear and lead to colour distortion, which make them unusable for downstream computer vision tasks (Fig.~\ref{fig:teaser}). This calls for greater exploration to enhance the image illumination for better performance on downstream tasks.

In this work, we propose and introduce a new computer vision task called nighttime dehaze-enhancement. In this task, a nighttime hazed image is jointly dehazed and illumination-enhanced. This task can be understood as a combination of low-light image enhancement and nighttime image dehazing. Additionally, to motivate further research on this task, we introduce a new-large scale dataset called \textit{Reside-$\beta$ Night} dataset. Our dataset consists of 4122 nighttime hazed images generated from 2061 scenes, and 2061 corresponding ground truth images from the Reside-$\beta$ dataset. Each scene contributes 2 images to our dataset, varying in the atmospheric scattering coefficient $\beta$~\cite{Narasimhan2002}, which controls the amount of haze. Our dataset can not only serve as a benchmark for this task, but also for other related tasks such as single image dehazing, video dehazing, and low-light image enhancement.

Moreover, we also propose a new deep learning-based algorithm called NDENet to solve nighttime image dehazing. Our network is based on Retinex theory, where we aim to improve the image quality by processing the illumination and reflectance. NDENet jointly performs the task of haze removal and light enhancement in an interrelated and end-to-end manner. We argue that by jointly solving the two tasks in an end-to-end fashion, we can leverage the complementary nature of the two tasks to produce high quality and clear images.
We also propose and evaluate several baseline algorithms using state-of-the-art algorithms for single image dehazing and low-light image enhancement. Experimental results clearly demonstrate the effectiveness of performing dehazing and enhancement together. The dataset and code will be made public upon acceptance.

The key contributions of our paper are as follows:
\begin{itemize}
    \item We introduce a new computer vision task called Nighttime Dehaze-Enhancement.
    \item We create the first large-scale dataset for this task, consisting of 4122 images along with 2061 clear ground truths.
    \item We propose a novel approach called NDENet for simultaneously dehazing and illuminating a scene. 
\end{itemize}

\section{Related Work}
\bfsection{Single Image Dehazing}
Most traditional single image dehazing methods build upon the Atmospheric Scattering model~\cite{Narasimhan2002}.  Under this optical model, the hazy image ($I$) is decomposed into transmission maps ($t$), global atmospheric illumination ($A$), and clear image ($J$) using Eq.~\ref{eq:atmo-scatter-model}. Several dehazing methods~\cite{invhaze2014, multiscale, jointtrans, li2017allinone} use the scattering model.

\begin{equation}
    J(z) = \frac{I(z) - A(1 - t(z))}{t(z)}
    \label{eq:atmo-scatter-model}
\end{equation}

Approaches following the atmospheric scattering model can be classified under two categories: \textit{prior-based} and \textit{learning-based}. The prior-based approach estimates the transmission maps and global atmospheric illumination using image priors. Dark Channel Prior~\cite{he2010dcp} estimated the transmission map based on empirical evidence that in an RGB image, one of the three colour channels is of significantly lesser intensity than the other two. 
Zhu \textit{et al.}~\cite{zhu2015cap} proposed the Colour Attenuation Prior (CAP), which performs dehazing based on the correlation of haze with the difference between brightness and saturation at each patch. Non-Local Colour Prior~\cite{berman2016non}, Change of Detail~\cite{li2015cod} and Colour Ellipsoid~\cite{bui2017colorellipsoid} are few other prior-based methods.

Prior-based methods fail to generalize to challenging environments, such as nighttime images, and learning-based methods attempt to learn the transmission map from the training data. Some methods estimate the transmission map (following the atmospheric scattering model) while others learn an end-to-end (RGB-to-RGB) dehazed image. Cai \textit{et al.} introduced DehazeNet~\cite{cai2016dehazenet}, one of the first RGB-to-RGB learning-based method for single image dehazing. Zhang \textit{et al.} proposed DCPDN~\cite{zhang2018dcpdn}, which used a GAN to estimate the transmission map and the atmospheric light. Cycle-Dehaze~\cite{engin2018cycledehaze} used a cycle-GAN to perform dehazing. Qin \textit{et al.} proposed FFA-Net~\cite{qin2020ffa} which performs RGB-to-RGB dehazing using Feature Attention (FA). BPPNet~\cite{singh2020bppnet} introduced Pyramid Convolution (PyCon) for dehazing using multiple input scales. The proposed nighttime dehaze-enhancement not only requires successful dehazing, but also enhancement of the image to make it usable for downstream tasks.

\bfsection{Nighttime dehazing}
Nighttime dehazing has recently gained prominence in literature, as daytime dehazing approaches often fail to generalize over nighttime scenes. Pei and Lee~\cite{pei2012nighttime} proposed a colour transfer based model augmented with dark channel prior and bilateral filter for haze removal. Li \textit{et al.}~\cite{li2015nighttime} solved non-uniform lighting by incorporating a glow factor alongside the atmospheric map and transmission estimation. Ancuti \textit{et al.}~\cite{ancuti2016night} computed patch-wise airlight components to deal with non-uniform lighting. Park \textit{et al.}~\cite{park2016nighttime} alleviated the haze and glow effect by estimating atmospheric light and transmission map using a weighted entropy. Zhang \textit{et al.}~\cite{zhang2017fast} targeted varying ambient illumination conditions by introducing a reflectance prior (MRP). OS-MRP~\cite{zhang2020nighttime} addressed the problem of colour correction and haze removal sequentially.

To reduce colour distortion and halo effects during nighttime dehazing, Yu \textit{et al.}~\cite{yu2019nighttime} proposed the estimation of transmission map using a pixel-wise alpha blending method. Lou \textit{et al.}~\cite{lou2020integrating} approached halo-free nighttime dehazing by using a linear learning model and a colour-dependent MRP. SIDE~\cite{he2020side} solved both night light enhancement and haze removal in a sequential manner by considering illumination and halo effects. STN~\cite{tang2021nighttime} introduced a structure-texture-noise decomposition model to consider only noise-free features for dehazing. Tang \textit{et al.}~\cite{tang2021nighttime} propose the use of Taylor series expansion for estimating the point-wise transmission map along with Retinex theory. Feng \textit{et al.}~\cite{feng2020learning} used an autoencoder to  estimate the transmission map and a guided filtering method to obtain ambient illumination. Nighttime dehazing approaches only focus on removing haze from nighttime images but do not enhance the image illumination, making it different from our task.

\bfsection{Low-light Image Enhancement}
The earliest illumination or contrast enhancement techniques utilized various conventional computer vision operations like increasing the brightness, saturation or dynamic range of the image, and Gamma Correction (GC). Later, histogram-based approaches~\cite{kim1997bihist, chen2003bihist, ibrahim2007histeq, singh2015histeq} were introduced, which were far more successful. The Retinex theory~\cite{land1977retinex} has also been leveraged for enhancement~\cite{park2017llretinex, li2018llretinex}, through which the image is decomposed into its illumination and reflectance components, following which the illumination map is enhanced. Recently, deep learning has also been used for this task~\cite{zhang2019kindling, Chen2018DeepRetinex, zhang2021kindplus, guo2020zero}. Low-light image enhancement does not involve removing haze or noise from the scenes and only focuses on increasing illumination.
\section{Reside-$\beta$ Night}
Since none of the existing benchmarks are suitable for the task of nighttime dehaze-enhancement, we create a large-scale benchmark for evaluation of algorithms on this task. This benchmark must satisfy several criteria. First, it must consist of nighttime hazed and corresponding dehazed image pairs. Second, since the nighttime hazed image is synthetic, it must be created by some non-trivial method consisting of challenging patterns of haze and varying degrees of illumination. Finally, the synthetic image must also appear realistic, so that the algorithms trained on this dataset can be used in real-world applications.

Keeping the criteria mentioned above in mind, we introduce a new dataset called the \textit{Reside-$\beta$ Night} dataset. We construct our dataset using hazed and dehazed image pairs from the Reside-$\beta$ dataset~\cite{li2019benchmarking}. 

The dataset is generated from the Reside-$\beta$ Outdoor Training Set (OTS) dataset \cite{li2019benchmarking} which consists of 2061 scenes. Each scene contains 35 hazed images and one corresponding clear image, having a total of 72,135 hazed images and 2061 clear images. Each of the 35 hazed images from every scene is synthetically generated by varying $A$ (global atmospheric light) and $\beta$ (scattering coefficient of atmosphere) parameters associated with the atmospheric scattering model. We select a subset of images from Reside-$\beta$ OTS dataset which have a parameter of $A=1$ and $\beta \in \{0.08, 0.16\}$. The resulting 4122 hazed images are split in $3:1$ ratio into training and test sets, containing 1541 and 520 images respectively. Next, the hazed images are transformed into HSV colour space, where the images are darkened. Finally, a gamma correction is performed to reduce illumination. This serves as our nighttime dehaze-enhancement dataset. 

\textit{Reside-$\beta$ Night} is the first large-scale dataset for nighttime dehaze-enhancement. This benchmark could also prove useful for similar tasks such as single image dehazing, video image dehazing, and low-light image enhancement. We believe our dataset will serve as a useful benchmark for various dehazing tasks, thereby enabling stronger performance on downstream scene understanding tasks.
\section{NDENet}
Our new network for nighttime dehaze-enhancement is illustrated in Fig.~\ref{fig:network}. The proposed network consists of three stages: 1) Decomposition, 2) Illumination Enhancement, and 3) Dehazing. The decomposition module is inspired from Retinex theory, where the input image is decomposed into corresponding reflectance and illumination components. The illumination enhancement module brightens the illumination map in a structurally aware manner. Finally, the dehazing module eliminates noise and haze. We explain our network in detail in the following sections.

\input{fig-tex/network}

\subsection{Decomposition}

Retinex theory aims to explain the colour perception during human vision. Under this theory, an image is assumed to consist of illumination and reflectance components. Concretely, the decomposition of an image $S$ is given as 

\begin{equation}
    \mathit{S} = \mathit{I} \circ \mathit{R}
    \label{eq:retinex-dec}
\end{equation}

where $I$ and $R$ denote the illumination and reflectance respectively, and $\circ$ represents the element-wise product.  Illumination indicates the lightness in the image. Reflectance denotes the intrinsic characteristic of objects in the image which lead to the sensation of colour. The illumination and reflectance components are independent to each other.

However, Retinex decomposition is an ill-posed problem~\cite{Chen2018DeepRetinex} and requires handcrafted tuning, which fails to generalize to challenging low-light conditions. Therefore, we approximate the Retinex decomposition using a convolutional neural network (CNN).

We model the decomposition CNN using a U-Net architecture~\cite{ronneberger2015unet}. Given an input RGB image, the decomposition network outputs the corresponding illumination and reflectance components for the image. During training, the network decomposes the nighttime haze and corresponding dehazed images to their illumination and reflectance components. However, since ground-truths are not available for illumination and reflectance, the decomposition is guided using loss functions to impose several constraints. Three loss functions are used to train the decomposition network: reconstruction loss ($L_{recon}$), reflectance similarity loss ($L_{rs}$), and illumination smoothness loss ($L_{is}$).


The decomposition network decomposes the nighttime hazed image $S_N$ and dehazed image $S_D$ to illumination ($I_N, I_D$) and reflectance components ($R_N, R_D$). To ensure that the learnt decomposition follows the Retinex theory, we use the following reconstruction loss:

\begin{equation}
    \mathit{L_{decom}} = \sum_{i \in \{D, N\}} \sum_{j \in \{D, N\}} \lambda_{ij} \lVert R_i \circ I_j - S_j \rVert
    \label{eq:decom-loss}
\end{equation}

where $\lambda_{ij}$ denotes the weight assigned to the reconstruction combination. Since reflectance is an intrinsic property of objects in the image, $R_D$ and $R_N$ must be similar while the illumination varies. To ensure that the learnt reflectances maintain this property, a given image $S_i$ is reconstructed using it's illumination map $I_i$ and \textit{both} reflectance maps $R_D$ and $R_N$. In other words, the reflectance for the hazed image is used to reconstruct the hazed image as well as the dehazed image, thereby ensuring that the reflectances $R_D$ and $R_N$ are similar.

In addition, a direct reflectance similarity constraint ensures that the reflectances $R_D$ and $R_N$ are similar by minimizing the following loss function:

\begin{equation}
    \mathit{L_{rs}} = \lVert R_N - R_D \rVert
    \label{eq:refl-loss}
\end{equation}

The learnt illumination must be locally consistent and preserve the structure to maintain smooth texture details~\cite{guo2016lime}. Therefore, we use the illumination smoothness loss as in~\cite{Chen2018DeepRetinex}. The Total variation minimization (TV)~\cite{chan2011augmented} is structure-agnostic, and therefore, is weighted using the gradient of reflectance to introduce texture and boundary context. This illumination smoothness loss is given as:

\begin{equation}
    \mathit{L_{is}} = \sum_{i \in \{D, N\}} \lVert \nabla I_i \circ \exp{(- \lambda _{s} \nabla R_i)} \rVert 
    \label{eq:illu-smooth-loss}
\end{equation}

where $\lambda _{s}$ denotes the weight assigned to the structure-awareness from the reflectance gradient. At the boundaries, the gradient is steep which introduces discontinuity in the learnt illumination.

\subsection{Illumination Enhancement}

Several handcrafted approaches have been explored to adjust the lightness of low-light images. However, these approaches fail to generalize well for images captured across challenging conditions such as dim lighting and low colour contrast. Therefore, we use a convolutional neural network (CNN) to perform brightness adjustment and enhance illumination.


The input to the network is the Retinex decomposition for the nighttime hazed image. In other words, the network is fed the illumination and reflectance components of the nighttime hazed image. The output of the enhancement network is the brightened illumination map. The reflectance is also used as an input so that the network can learn to adjust the illumination of the image in a context-aware manner. Concretely, this enables the network to dynamically enhance the brightness of specific objects not only using shape information from the illumination map, but also using the colour and texture information from the reflectance map. We use 11 convolution layers in the enhancement network with residual connections at each stage.


\subsection{Dehazing}
Greater noise is often encountered in nighttime images. While illumination enhancement adjusts the image brightness, it also results in haze amplification. Hence, the reflectance component needs to be dehazed to eliminate noise. Traditional daytime dehazing approaches use optical models with priors such as the dark channel prior, colour attenuation prior, etc. However, several of these priors do not hold for nighttime haze images due to the presence of several artificial sources of light~\cite{tang2021nighttime}. Therefore, these methods struggle to dehaze nighttime images and often lead to unnatural outputs consisting of drastic colour distortion and halo artifacts. Therefore, we apply a learning-based algorithm to perform reflectance dehazing.

We follow a state-of-the-art single image dehazing network, namely, Densely Connected Pyramid Dehazing Network (DCPDN)~\cite{zhang2018dcpdn}, for this purpose. Specifically, we use the transmission map estimation network with the objective of reflectance dehazing. Our network consists of an encoder-decoder structure with densely connected blocks to improve feature information flow for robust feature learning. Additionally, a multi-level pyramid pooling is used to capture context from multiple scales. The decoder consists of bottleneck blocks with residual connections, with transition blocks introduced to restore the resolution lost due to upsampling operations.

The first 3 Dense blocks of DenseNet-121~\cite{huang2017densenet} are used as the encoder network, with 4 pooling sizes of 1/32, 1/16, 1/8, 1/4 used in the pyramid pooling layer. Each Bottleneck layer consists of two \textit{BatchNorm-Conv-ReLU-Dropout} mini-blocks.

\subsection{Reconstruction}
The adjusted illumination and dehazed reflectance components are used to reconstruct the dehazed image using Eq.~\ref{eq:retinex-dec}. The image reconstruction loss is formulated by minimizing the mean-squared error between the reconstructed image $S_Y$ and target dehazed image $S_D$ . In addition, we also use self-supervision to ensure robust decomposition by minimizing the mean-squared error between the respective illuminations ($I_D$ and $I_Y$) and reflectances ($R_D$ and $R_Y$). This reconstruction loss $L_{MSE}$ is given as:

\begin{equation}
    \mathit{L_{MSE}} = \sum_{F \in \{I, R, S\}} \lambda _F \lVert F_Y - F_D \rVert ^2
    \label{eq:recon-mse-loss}
\end{equation}

where $\lambda_F$ denotes the weight assigned to each loss function.

Moreover, to ensure the perceptual similarity between $S_Y$ and $S_D$, we use the VGG-based edge-preserving loss~\cite{simonyan2014vgg, Johnson2016Perceptual}. The edge-preserving loss minimizes the feature-level difference between the two images using a pre-trained VGG network $\phi$. This difference is computed at each layer $i$ of the VGG feature extractor. The edge-preserving loss is formulated as:

\begin{equation}
    \mathit{L_{VGG}} = \lambda _{\phi} \sum_{i} w_i \lVert \phi (\mathit{S_Y})_i - \phi (\mathit{S_D})_i \rVert ^2
    \label{eq:recon-vgg-loss}
\end{equation}

where $\lambda _{\phi}$ denotes the weight assigned to the edge-preserving loss, and $w_i$ denotes the weight of each layer towards the edge-preserving loss. The final reconstruction loss is:

\begin{equation}
    \mathit{L_{recon}} = \mathit{L_{VGG}} + \mathit{L_{MSE}}
    \label{eq:recon-loss}
\end{equation}
\section{Experiments}
In this section, we demonstrate the effectiveness of the proposed algorithm by conducting several qualitative and quantitative comparisons with state-of-the-art methods for single image dehazing and low-light enhancement on the Reside-$\beta$ Night dataset.

\input{fig-tex/results}
\input{fig-tex/night_results}

\subsection{Implementation Details}
Following the training protocol in~\cite{Chen2018DeepRetinex}, the network is trained in two stages. In the first stage, we train the Retinex decomposition network individually. The input images and corresponding ground truths in the decomposition UNet are cropped to a size of $256 \times 256$. We use 32 channels in the first layer of the UNet to maintain a small memory footprint for the decomposition network. In the second stage, the weights of the decomposition network are not updated, and the remaining network is trained.

While training the decomposition module, the network is fed nighttime hazed images as well as the corresponding dehazed images with shared weights, and it outputs reflectance and illumination for both. However, for the second stage, the entire network is used but the decomposition part of the network is frozen. Only the nighttime hazed image is fed to the decomposition module, and the corresponding reflectance and illumination are subsequently enhanced by the enhancement module in the second stage.


In both training stages, we use an Adam optimizer with a learning rate of $2.5e-4$ to train the networks. The first and second stages are trained for 55 and 25 epochs, respectively, with a batch size of 2. An Nvidia T4 GPU is used for to train our network. We also apply data augmentation methods (random resizing, cropping) during training.

For $L_{decom}$ (Eq.~\ref{eq:decom-loss}), we use $\lambda _{DD} = \lambda _{NN} = 1$, $\lambda _{ND} = 0.01$ and $\lambda _{DN} = 0.001$. For the reconstruction MSE loss $L_{MSE}$ (Eq.~\ref{eq:recon-mse-loss}), we set $\lambda _S = 1$, $\lambda _I = 0.01$, and $\lambda _R = 0.05$. For the edge-preserving loss (Eq.~\ref{eq:recon-vgg-loss}), we use $\lambda _\phi = 1$ with weights of $[8, 4, 2, 1]$ for each VGG layer.

\subsection{Results}

\input{tab-tex/ll-dh-comparison}
\input{tab-tex/dh-ll-comparison}

We evaluate our method against four state-of-the-art approaches on single image dehazing and low-light image enhancement. For single image dehazing, we evaluate DCPDN~\cite{zhang2018dcpdn} and FFANet~\cite{qin2020ffa}, and for low-light image enhancement, we evaluate Zero-DCE~\cite{guo2020zero} and KinD++~\cite{zhang2021kindplus}. However, we note that direct inference through these networks is not a fair comparison, as they are trained with different objectives. Therefore, we cascade the single image dehazing and low-light enhancement in different orders to compare our approach. In Table~\ref{table:comparison-ll-dh}, the image is first enhanced using the enhancement network and then dehazed, while in Table~\ref{table:comparison-dh-ll}, the order is reversed, i.e., the image is first dehazed and then enhanced. In Table~\ref{table:comparison-night}, our method is compared against two nighttime dehazing methods: DNU~\cite{santra2016dnu} and MRP~\cite{jing2017mrp}.

We also demonstrate the qualitative performance of NDENet in Fig~\ref{fig:results} and compare it with the 4 best-performing combinations of dehazing and enhancement networks. We note that performing low-light enhancement first followed by dehazing results in a mute tone but hazy image. Performing Dehazing first produces more colour contrast, yet the image texture and sharpness do not remain consistent throughout the image. We observe that FFANet -- Zero-DCE performs well in dehazing but does not increase the illumination, while FFANet -- KinD++ is able to recover the original brightness of the image while leading to unnatural sharpness and colour contrast. In all compared methods, regions of the image having higher haze density still appear hazy. However, our approach is able to dehaze the scene successfully, while maintaining texture consistency throughout.

We also evaluate our network against two open-source nighttime dehazing methods -- DNU~\cite{santra2016dnu} and MRP~\cite{jing2017mrp} in Table~\ref{table:comparison-night-ll}. We use two settings: direct nighttime dehazing and cascaded with low-light enhancement networks. We also compare the qualitative performance of our network with two direct nighttime dehazing methods and the two best performing cascade methods in Fig~\ref{fig:results_night}.

Clearly, the dehazed outputs from other approaches still contain noise, making them unusable for downstream tasks. The nighttime dehazing methods increase the contrast, evident in plain textures like sky and roads, while ruining the texture. The MRP -- Zero-DCE cascade results in a mute tone with observable noise and high saturation at the objects edges. The DNU -- Zero-DCE cascade results in high degree of noise with noticeable colour distortion.

\clearpage

\input{tab-tex/night-comparison}
\input{tab-tex/night-ll-comparison}

However, our method is able to maintain the colour and texture details well. Our method consistently outperforms all baselines, achieving an SSIM of 0.8962 and a PSNR of 26.25.

\section{Conclusion}
We introduce a new computer vision task named as nighttime dehaze-enhancement along with a new large-scale dataset called the Reside-$\beta$ Night dataset. This task combines nighttime image dehazing and low-light image enhancement, with an objective to jointly dehaze and enhance illumination. We also propose a new algorithm, NDENet, as a baseline algorithm with SSIM of 0.8962 and PSNR of 26.25, for future research on this task. We believe this task will motivate the research community to develop new ideas for scene understanding tasks, especially involving challenging nighttime scenes.


%





\ifCLASSOPTIONcaptionsoff
  \newpage
\fi



%

\bibliography{references}{}
\bibliographystyle{IEEEtran}




%








\end{document}

%% file: fig-tex/teaser.tex

\begin{figure}[t]
    \centering
    \setlength\tabcolsep{1.0pt}
    \begin{tabular}{ccc}
        \small
        (a) & \includegraphics[width=0.45\linewidth]{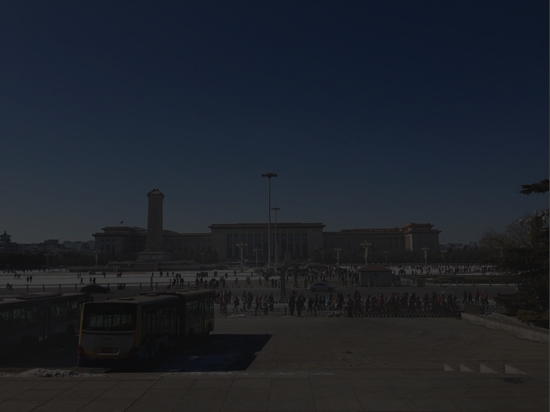} & 
        \includegraphics[width=0.45\linewidth]{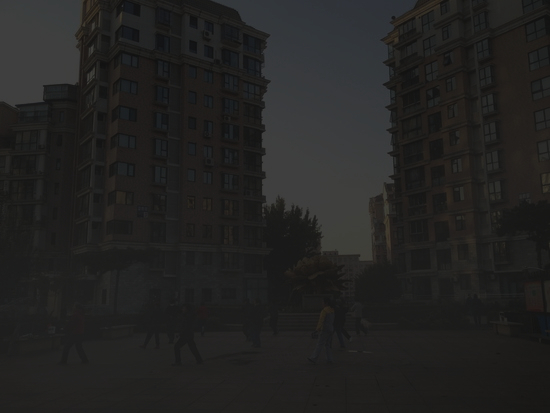}\\
        \small
        (b) & \includegraphics[width=0.45\linewidth]{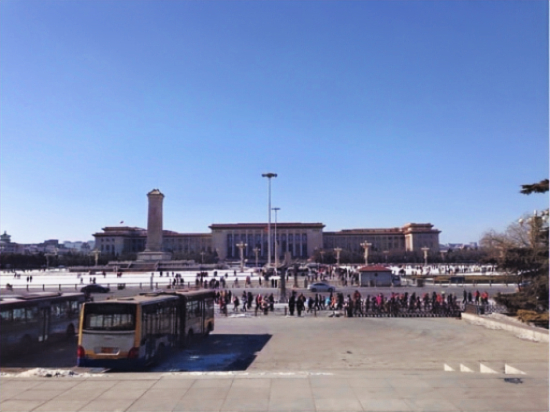} & 
        \includegraphics[width=0.45\linewidth]{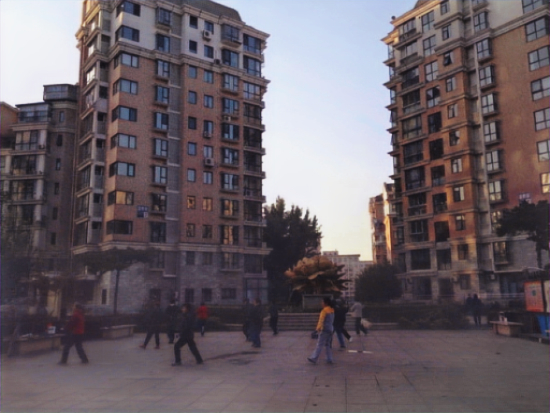}\\
        \small
        (c) & \includegraphics[width=0.45\linewidth]{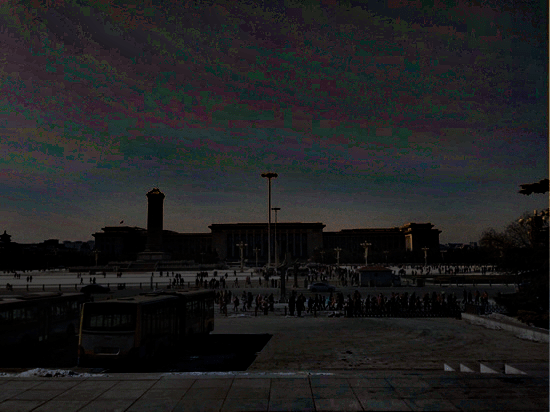} &
        \includegraphics[width=0.45\linewidth]{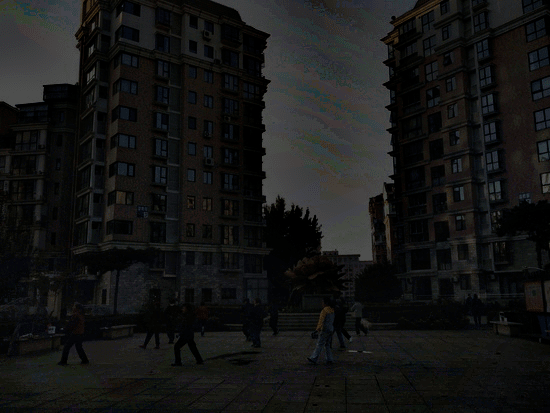}\\
    \end{tabular}
    \caption{The problem with nighttime dehazing methods. While the haze has been removed to an extent, the details are still hidden due to low illumination. (a) Nighttime hazy images, (b) Our method, (c) MRP~\cite{jing2017mrp}
    }
    \label{fig:teaser}
\end{figure}



%% file: fig-tex/network.tex
\begin{figure*}[t]
    \centering
    \includegraphics[width=\textwidth]{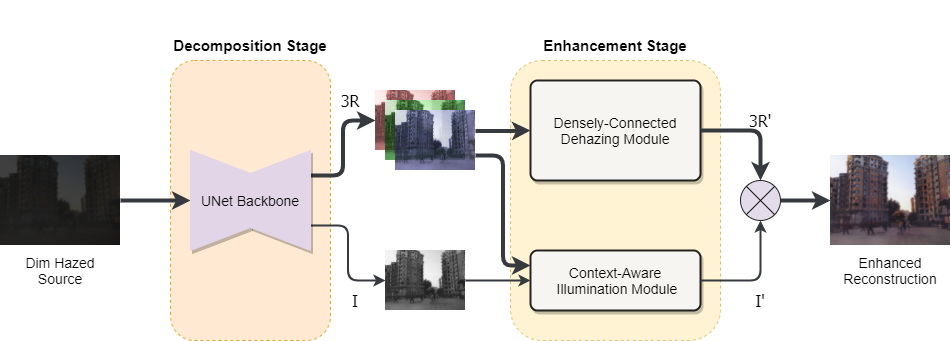}
    \caption{The proposed NDENet architecture.}
    \label{fig:network}
\end{figure*}

%% file: fig-tex/results.tex
\begin{figure*}[t]
    \centering
    \setlength\tabcolsep{1.0pt}
    \begin{tabular}{ccccc}
        \small
        \rot{Image} &
        \includegraphics[width=0.21\textwidth]{images/comparison/dark_1.jpg} & \includegraphics[width=0.21\textwidth]{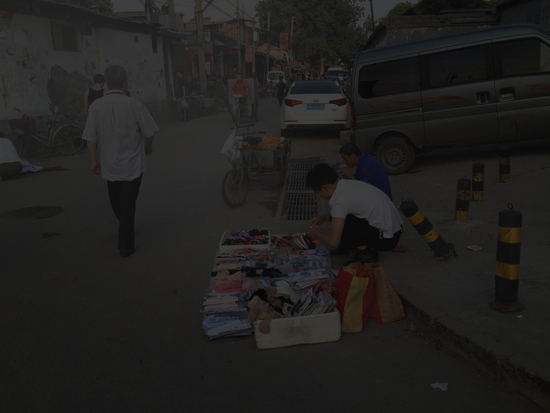} & \includegraphics[width=0.21\textwidth]{images/comparison_again/hazy_dark/2945_1_0.16.jpg} & \includegraphics[width=0.21\textwidth]{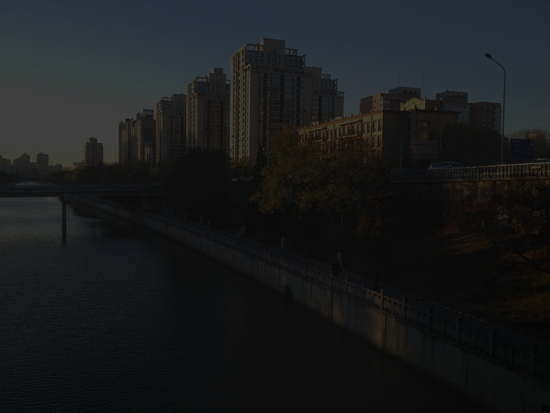}\\
        \small
        \rot{Ground truth} &
        \includegraphics[width=0.21\textwidth]{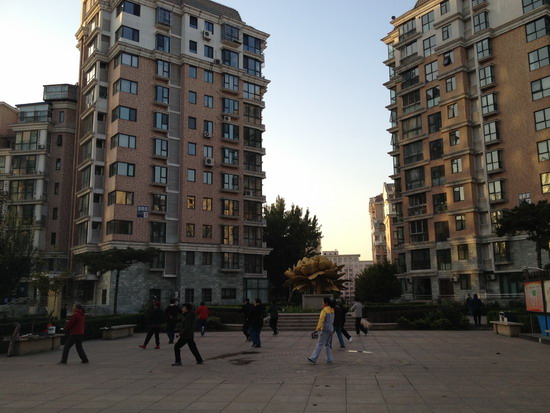} & \includegraphics[width=0.21\textwidth]{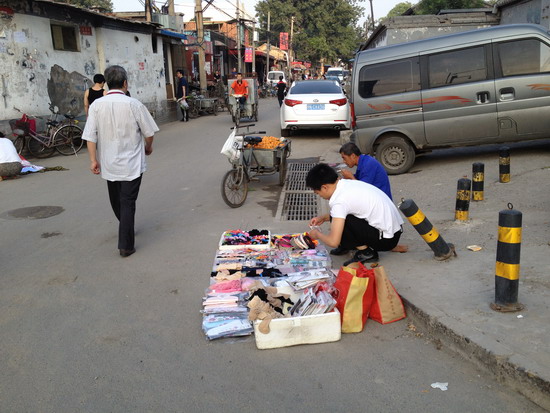} & \includegraphics[width=0.21\textwidth]{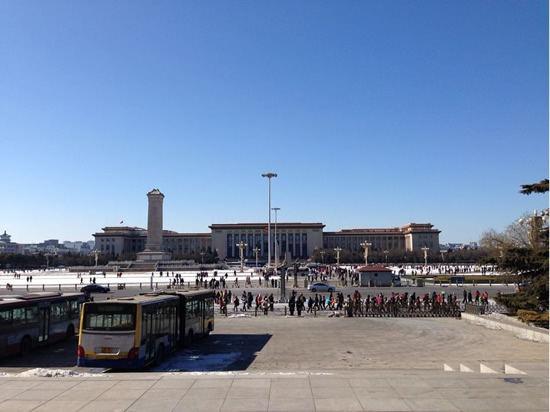} & \includegraphics[width=0.21\textwidth]{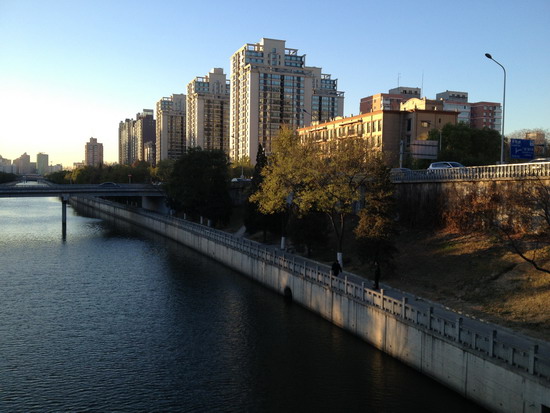}\\
        \small
        \rot{\textbf{NDENet (ours)}} &
        \includegraphics[width=0.21\textwidth]{images/comparison/out_1.jpg} & \includegraphics[width=0.21\textwidth]{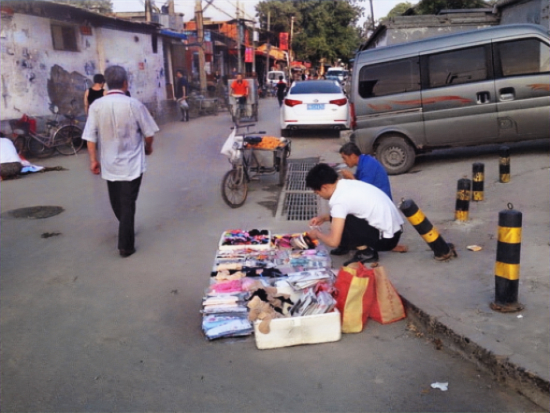} & \includegraphics[width=0.21\textwidth]{images/comparison_again/our/2945_1_0.16.jpg} & \includegraphics[width=0.21\textwidth]{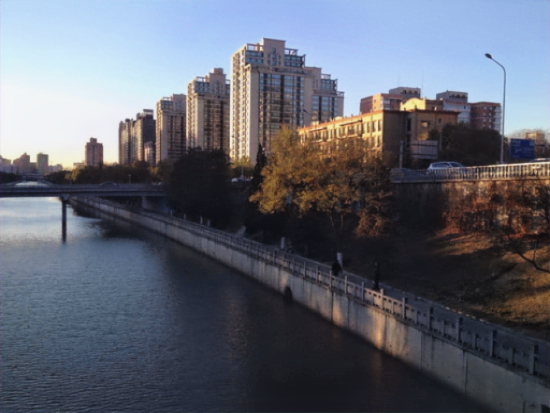}\\
        \small
        \rot{Zero-DCE -- DCPDN} &
        \includegraphics[width=0.21\textwidth]{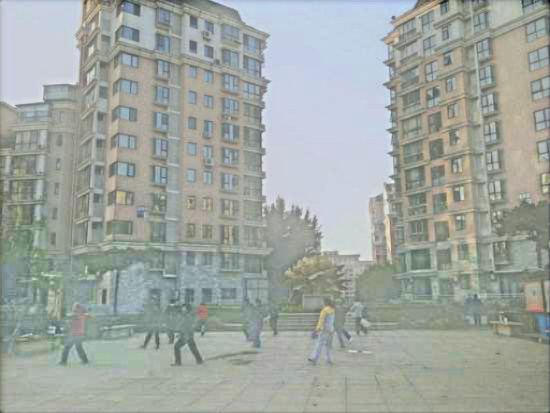} & \includegraphics[width=0.21\textwidth]{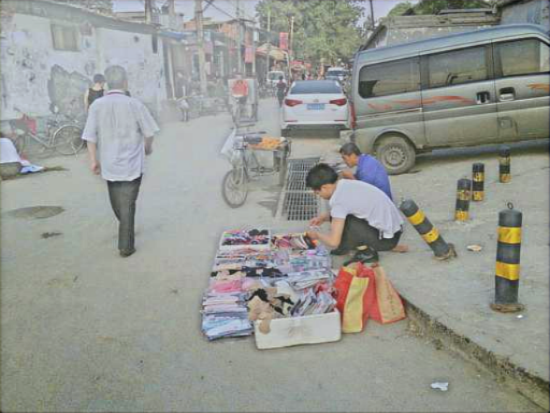} & \includegraphics[width=0.21\textwidth]{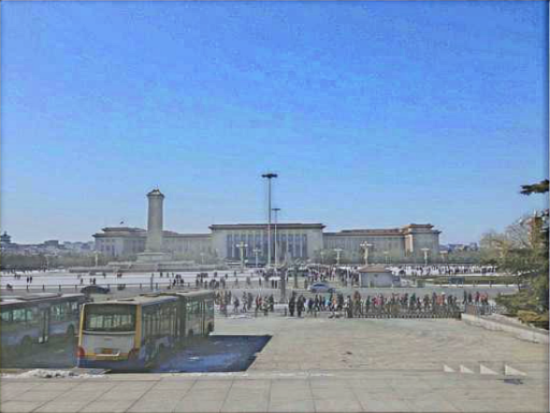} & \includegraphics[width=0.21\textwidth]{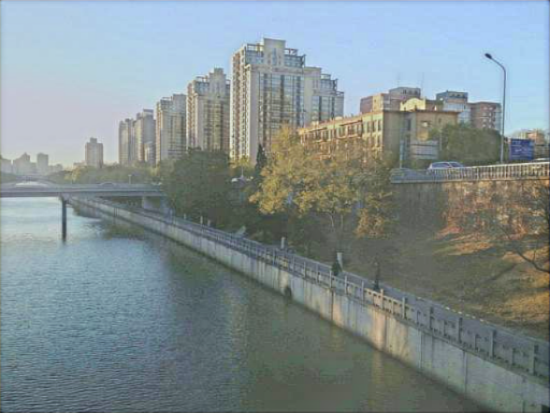}\\
        \small
        \rot{Zero-DCE -- FFANet} &
        \includegraphics[width=0.21\textwidth]{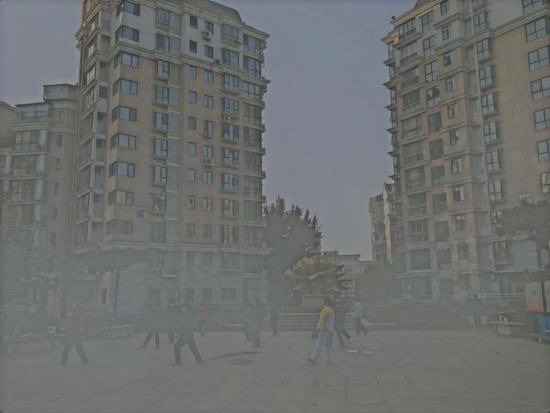} & \includegraphics[width=0.21\textwidth]{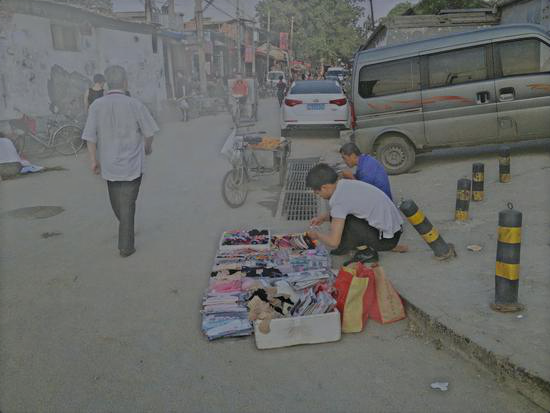} & \includegraphics[width=0.21\textwidth]{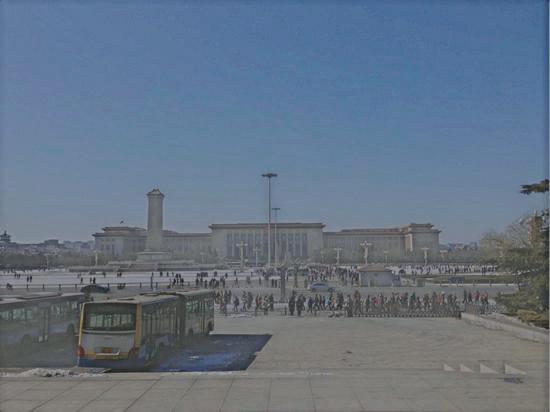} & \includegraphics[width=0.21\textwidth]{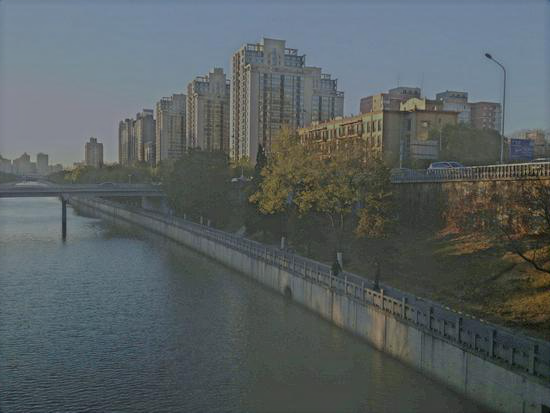}\\
        \small
        \rot{FFANet -- Zero-DCE} &
        \includegraphics[width=0.21\textwidth]{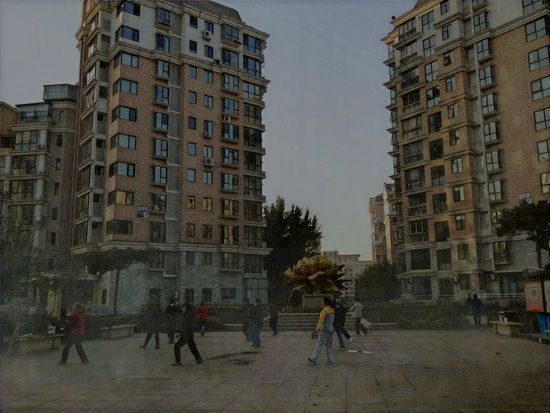} & \includegraphics[width=0.21\textwidth]{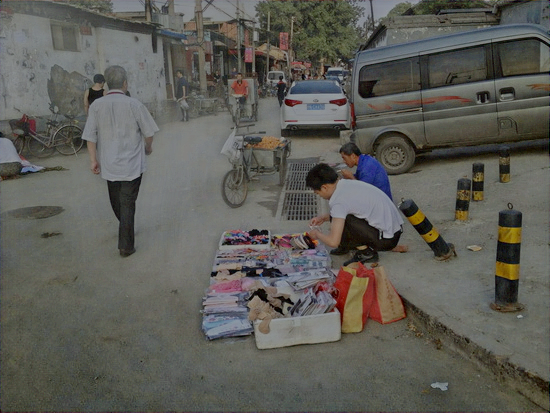} & \includegraphics[width=0.21\textwidth]{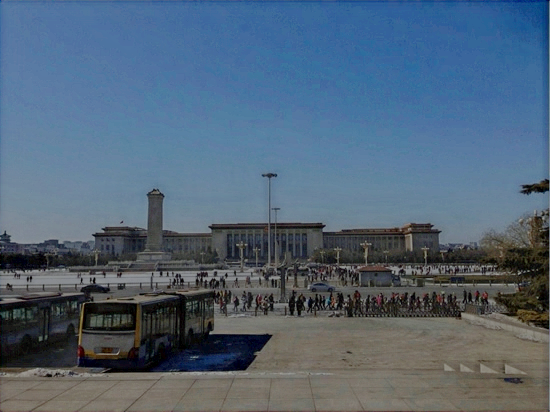} & \includegraphics[width=0.21\textwidth]{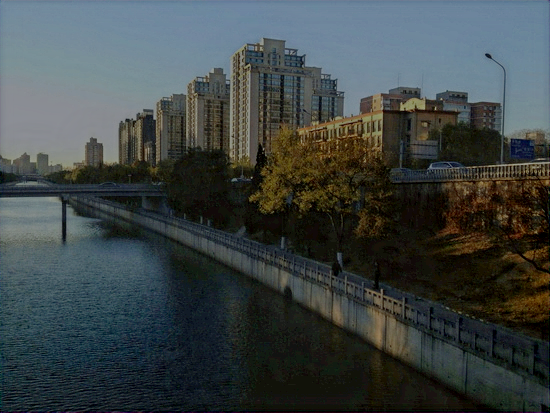}\\
        \small
        \rot{FFANet -- KinD++} &
        \includegraphics[width=0.21\textwidth]{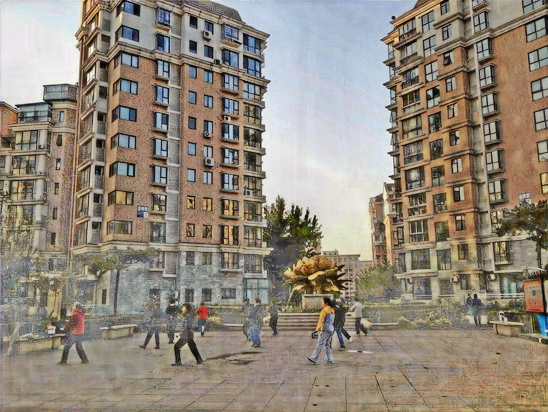} & \includegraphics[width=0.21\textwidth]{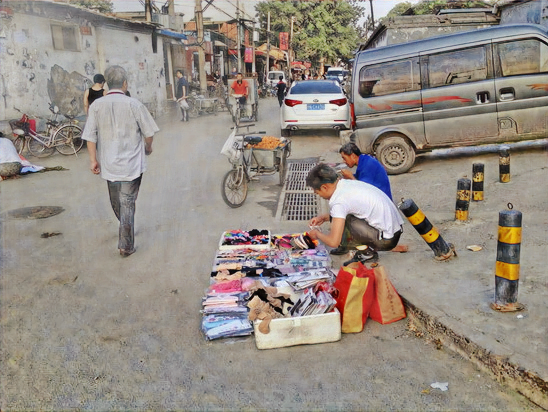} & \includegraphics[width=0.21\textwidth]{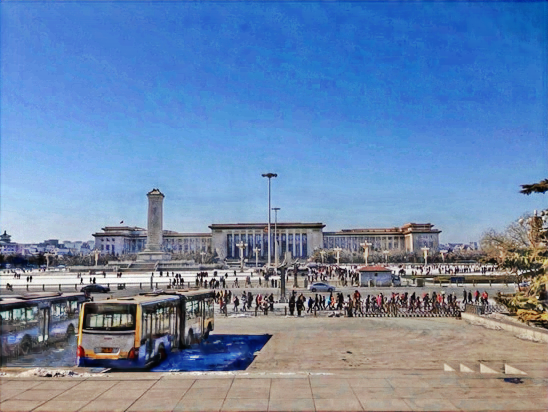} & \includegraphics[width=0.21\textwidth]{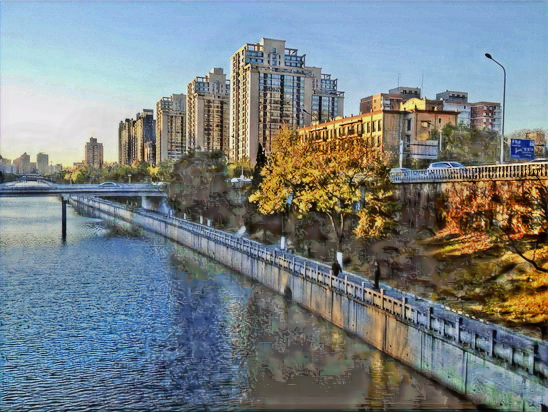}\\
    \end{tabular}

    \caption{We compare results obtained using our approach (third row) with other state-of-the-art methods cascaded together. Only the top-4 previous methods are showcased here. Clearly, our method is able to brighten images while maintaining details across hazy conditions.
    }
    \label{fig:results}
\end{figure*}

%% file: fig-tex/night_results.tex
\begin{figure*}[t]
    \centering
    \setlength\tabcolsep{1.0pt}
    \begin{tabular}{ccccc}
        \small
        \rot{Image} &
        \includegraphics[width=0.21\textwidth]{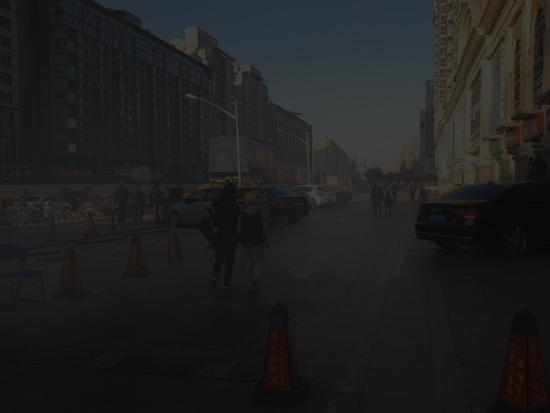} &
        \includegraphics[width=0.21\textwidth]{images/comparison_again/hazy_dark/something.jpg} &
        \includegraphics[width=0.21\textwidth]{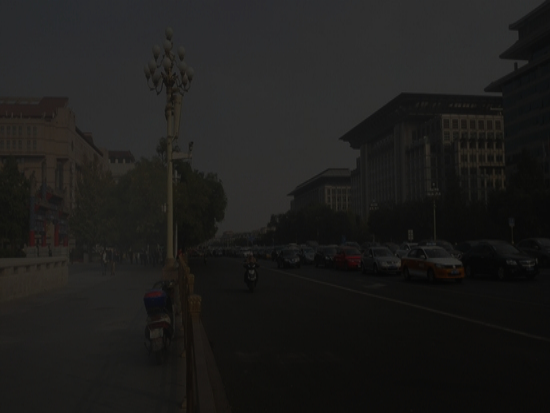} &
        \includegraphics[width=0.21\textwidth]{images/comparison_again/hazy_dark/2945_1_0.16.jpg}\\
        \small
        \rot{Ground truth} &
        \includegraphics[width=0.21\textwidth]{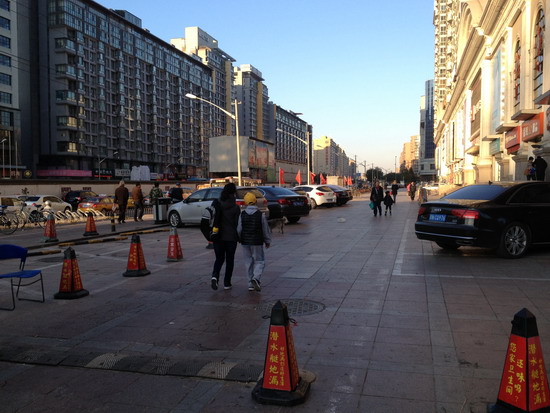} &
        \includegraphics[width=0.21\textwidth]{images/comparison_again/clear/something.jpg} &
        \includegraphics[width=0.21\textwidth]{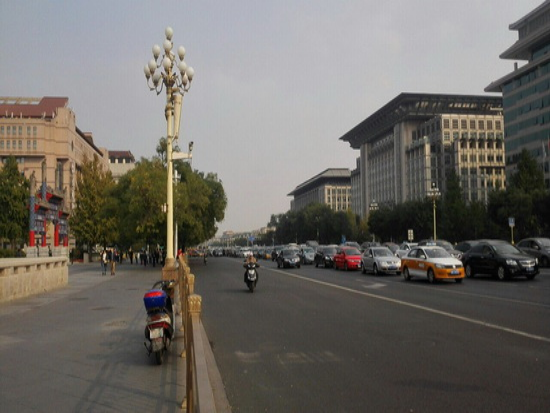} &
        \includegraphics[width=0.21\textwidth]{images/comparison_again/clear/2945.jpg}\\
        \small
        \rot{\textbf{NDENet (ours)}} &
        \includegraphics[width=0.21\textwidth]{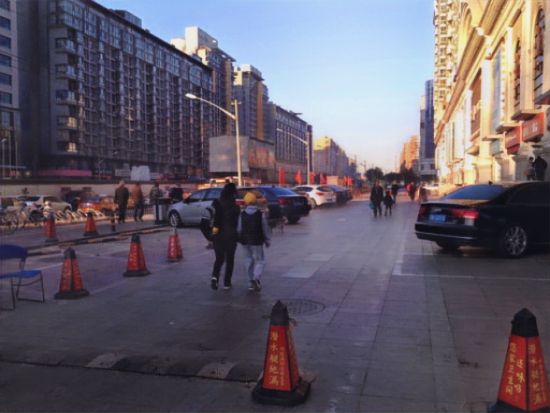} &
        \includegraphics[width=0.21\textwidth]{images/comparison_again/our/something.jpg} &
        \includegraphics[width=0.21\textwidth]{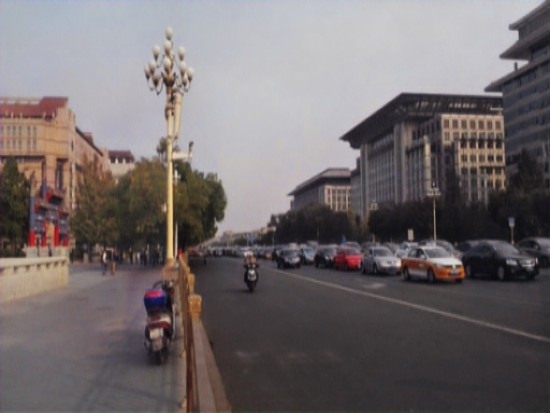} &
        \includegraphics[width=0.21\textwidth]{images/comparison_again/our/2945_1_0.16.jpg}\\
        \small
        \rot{MRP~\cite{jing2017mrp}} &
        \includegraphics[width=0.21\textwidth]{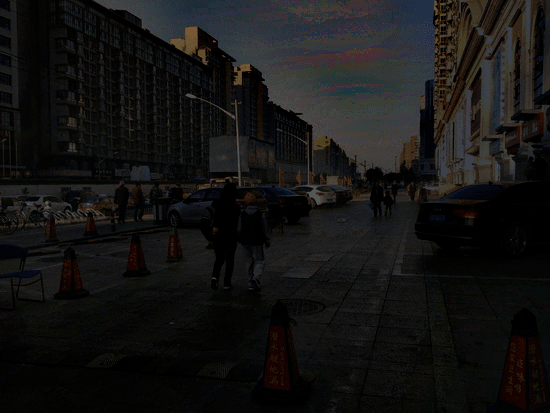} &
        \includegraphics[width=0.21\textwidth]{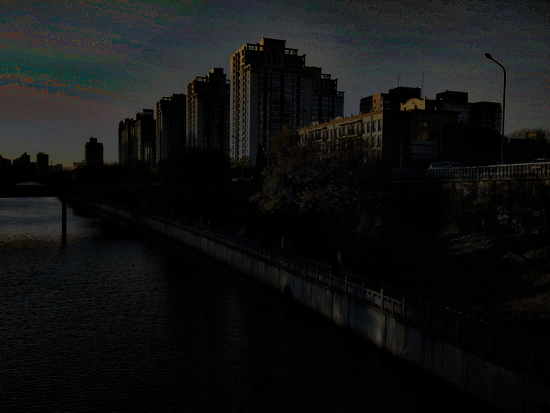} &
        \includegraphics[width=0.21\textwidth]{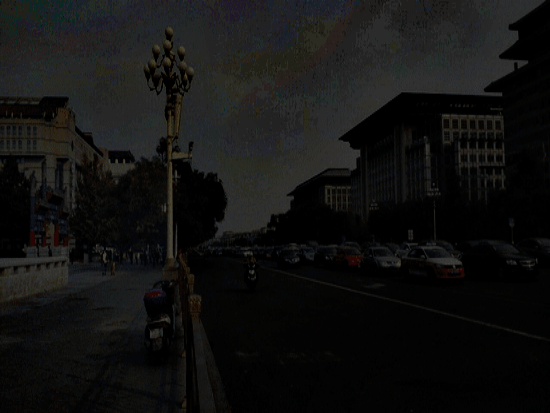} &
        \includegraphics[width=0.21\textwidth]{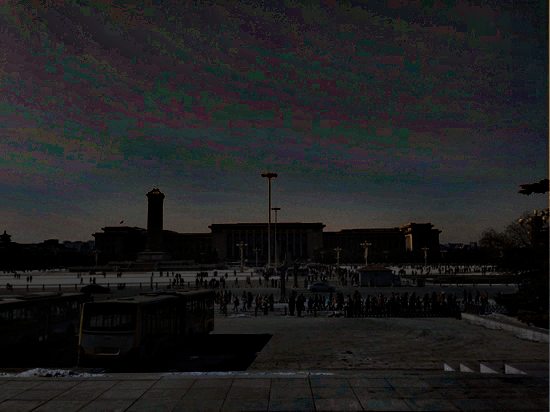}\\
        \small
        \rot{DNU~\cite{santra2016dnu}} &
        \includegraphics[width=0.21\textwidth]{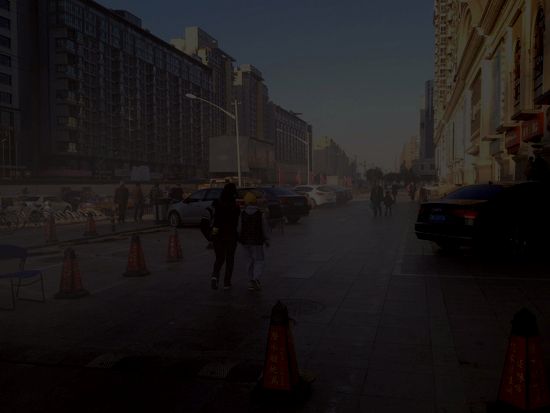} &
        \includegraphics[width=0.21\textwidth]{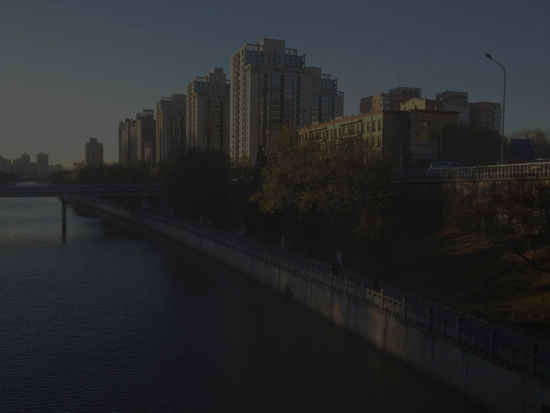} &
        \includegraphics[width=0.21\textwidth]{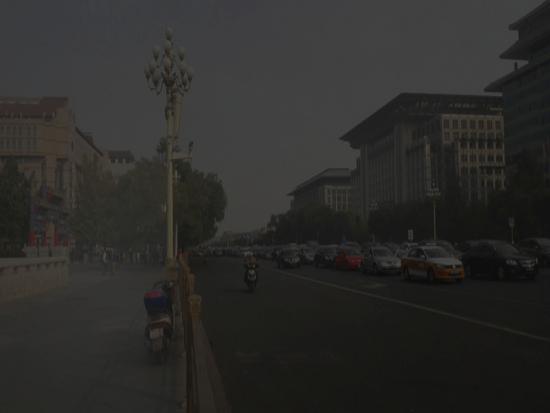} &
        \includegraphics[width=0.21\textwidth]{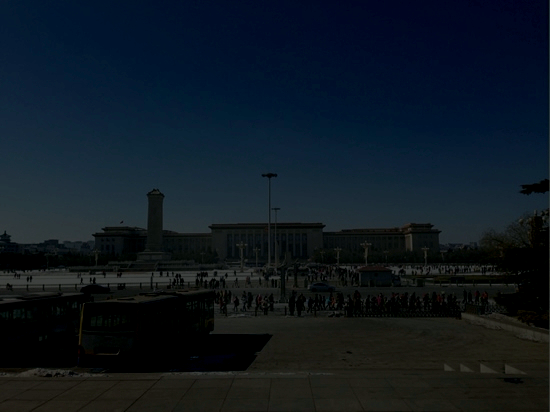}\\
        \small
        \rot{MRP -- Zero-DCE} &
        \includegraphics[width=0.21\textwidth]{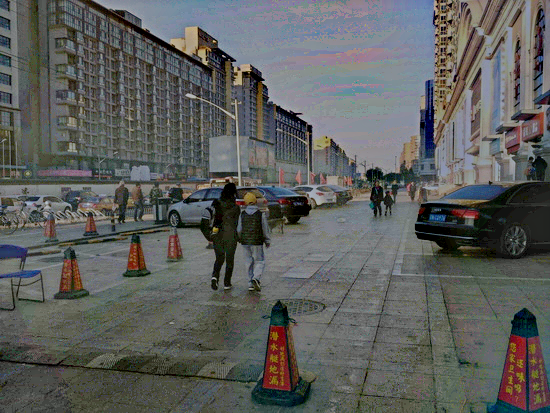} &
        \includegraphics[width=0.21\textwidth]{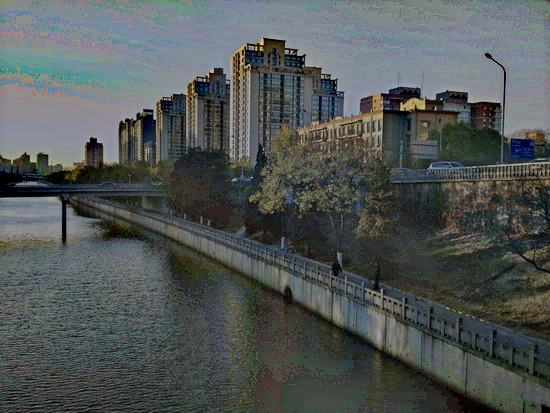} &
        \includegraphics[width=0.21\textwidth]{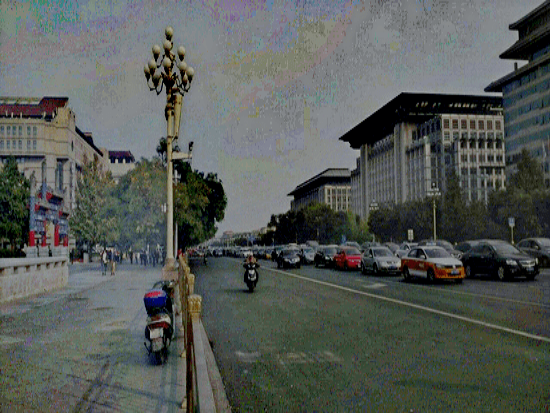} &
        \includegraphics[width=0.21\textwidth]{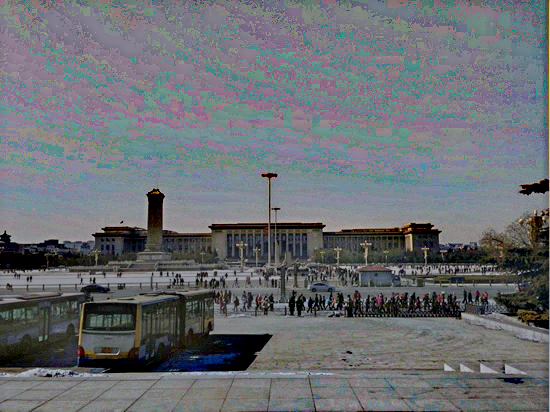}\\
        \small
        \rot{DNU -- Zero-DCE} &
        \includegraphics[width=0.21\textwidth]{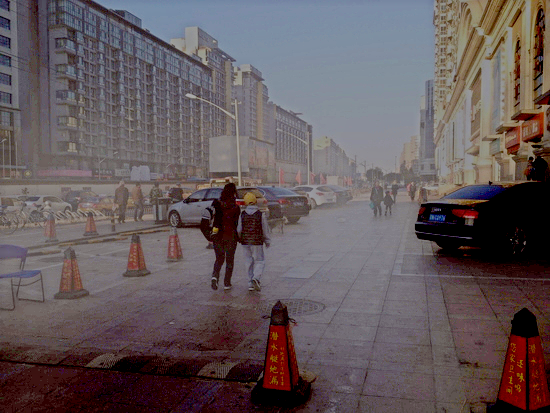} &
        \includegraphics[width=0.21\textwidth]{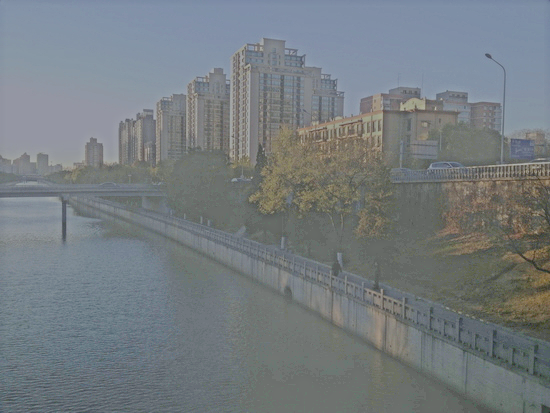} &
        \includegraphics[width=0.21\textwidth]{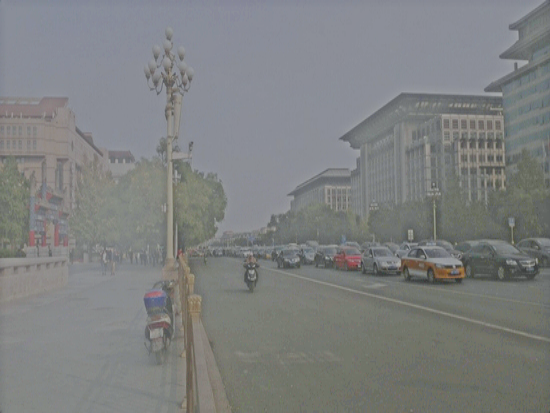} &
        \includegraphics[width=0.21\textwidth]{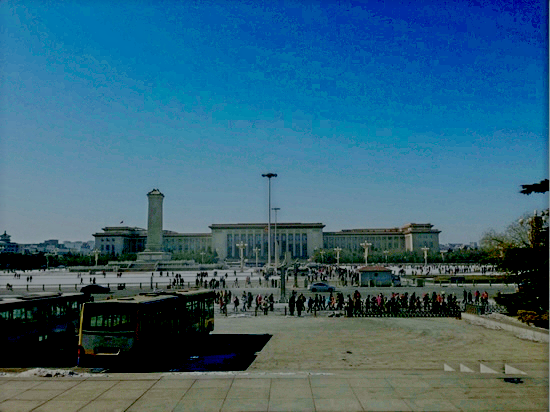}\\
    \end{tabular}

    \caption{We compare results obtained using our approach (third row) with other publicly available nighttime dehazing methods. Clearly, the previous night time methods struggle to generate images with visible details. Cascading enhancement networks (last two rows) distorts the image quality as well.
    }
    \label{fig:results_night}
\end{figure*}

%% file: tab-tex/ll-dh-comparison.tex
\begin{table}[t]
    \centering
    \begin{tabular}{c|c|c|c}
        \hline
        \textbf{Low-Light} & \textbf{Dehazing} & \textbf{SSIM} & \textbf{PSNR} \\
        \hline
        Zero-DCE & DCPDN & 0.7321 & 13.26 \\
        Zero-DCE & FFANet & 0.7993 & 15.16 \\
        KinD++ & DCPDN & 0.5007 & 9.63 \\
        KinD++ & FFANet & 0.5874 & 14.15 \\\hline
        \multicolumn{2}{c|}{\textbf{NDENet (ours)}} & \textbf{0.8962} & \textbf{26.25} \\
        \hline
    \end{tabular}
    \caption{Quantitative comparison of the proposed method with baselines. The image is first passed through a Low-Light Enhancement network followed by a Dehazing network.}
    \label{table:comparison-ll-dh}
\end{table}

%% file: tab-tex/dh-ll-comparison.tex
\begin{table}[t]
    \centering
    \begin{tabular}{c|c|c|c}
        \hline
        \textbf{Dehazing} & \textbf{Low-Light} & \textbf{SSIM} & \textbf{PSNR} \\
        \hline
        DCPDN & Zero-DCE & 0.6345 & 9.61 \\
        DCPDN & KinD++ & 0.6774 & 11.72 \\
        FFANet & Zero-DCE & 0.8159 & 15.10 \\
        FFANet & KinD++ & 0.5639 & 15.16 \\\hline
        \multicolumn{2}{c|}{\textbf{NDENet (ours)}} & \textbf{0.8962} & \textbf{26.25} \\
        \hline
    \end{tabular}
    \caption{Quantitative comparison of the proposed method with baselines. The image is first passed through a Dehazing network followed by a Low-Light Enhancement network.}
    \label{table:comparison-dh-ll}
\end{table}

%% file: tab-tex/night-comparison.tex
\begin{table}[t]
    \centering
    \begin{tabular}{c|c|c}
        \hline
        \textbf{Nighttime Dehazing} & \textbf{SSIM} & \textbf{PSNR} \\
        \hline
        MRP~\cite{jing2017mrp} & 0.3165 & 7.66 \\
        DNU~\cite{santra2016dnu} & 0.3825 & 8.15 \\
        \textbf{NDENet (ours)} & \textbf{0.8962} & \textbf{26.25} \\
        \hline
    \end{tabular}
    \caption{Quantitative comparison of the proposed method with nighttime dehazing algorithms.}
    \label{table:comparison-night}
\end{table}

%% file: tab-tex/night-ll-comparison.tex
\begin{table}[t]
    \centering
    \begin{tabular}{c|c|c|c}
        \hline
        \textbf{Nighttime Dehazing} & \textbf{Low-Light} & \textbf{SSIM} & \textbf{PSNR} \\
        \hline
        MRP & Zero-DCE & 0.6871 & 14.56 \\
        MRP & KinD++ & 0.5116 & 13.39 \\
        DNU & Zero-DCE & 0.7313 & 14.79 \\
        DNU & KinD++ & 0.5794 & 13.83 \\
        \hline
        \multicolumn{2}{c|}{\textbf{NDENet (ours)}} & \textbf{0.8962} & \textbf{26.25} \\
        \hline
    \end{tabular}
    \caption{Quantitative comparison of the proposed method with nighttime dehazing algorithms cascaded with a Low-Light Enhancement network.}
    \label{table:comparison-night-ll}
\end{table}

%% file: journal.bbl
\begin{thebibliography}{10}
\providecommand{\url}[1]{#1}
\csname url@samestyle\endcsname
\providecommand{\newblock}{\relax}
\providecommand{\bibinfo}[2]{#2}
\providecommand{\BIBentrySTDinterwordspacing}{\spaceskip=0pt\relax}
\providecommand{\BIBentryALTinterwordstretchfactor}{4}
\providecommand{\BIBentryALTinterwordspacing}{\spaceskip=\fontdimen2\font plus
\BIBentryALTinterwordstretchfactor\fontdimen3\font minus
  \fontdimen4\font\relax}
\providecommand{\BIBforeignlanguage}[2]{{%
\expandafter\ifx\csname l@#1\endcsname\relax
\typeout{** WARNING: IEEEtran.bst: No hyphenation pattern has been}%
\typeout{** loaded for the language `#1'. Using the pattern for}%
\typeout{** the default language instead.}%
\else
\language=\csname l@#1\endcsname
\fi
#2}}
\providecommand{\BIBdecl}{\relax}
\BIBdecl

\bibitem{he2010dcp}
K.~He, J.~Sun, and X.~Tang, ``Single image haze removal using dark channel
  prior,'' \emph{IEEE transactions on pattern analysis and machine
  intelligence}, vol.~33, 08 2010.

\bibitem{jing2017mrp}
J.~Zhang, Y.~Cao, S.~Fang, Y.~Kang, and C.~W. Chen, ``Fast haze removal for
  nighttime image using maximum reflectance prior,'' in \emph{IEEE Conference
  on Computer Vision and Pattern Recognition (CVPR)}, 2017.

\bibitem{tang2021nighttime}
Q.~Tang, J.~Yang, X.~He, W.~Jia, Q.~Zhang, and H.~Liu, ``Nighttime image
  dehazing based on retinex and dark channel prior using taylor series
  expansion,'' \emph{Computer Vision and Image Understanding}, vol. 202, p.
  103086, 2021.

\bibitem{Narasimhan2002}
\BIBentryALTinterwordspacing
S.~G. Narasimhan and S.~K. Nayar, ``Vision and the atmosphere,''
  \emph{International Journal of Computer Vision}, vol.~48, no.~3, pp.
  233--254, Jul 2002. [Online]. Available:
  \url{https://doi.org/10.1023/A:1016328200723}
\BIBentrySTDinterwordspacing

\bibitem{invhaze2014}
K.~Tang, J.~Yang, and J.~Wang, ``Investigating haze-relevant features in a
  learning framework for image dehazing,'' in \emph{2014 IEEE Conference on
  Computer Vision and Pattern Recognition}, 2014, pp. 2995--3002.

\bibitem{multiscale}
W.~Ren, S.~Liu, H.~Zhang, J.~Pan, X.~Cao, and M.-H. Yang, ``Single image
  dehazing via multi-scale convolutional neural networks,'' in \emph{Computer
  Vision -- ECCV 2016}, B.~Leibe, J.~Matas, N.~Sebe, and M.~Welling, Eds.\hskip
  1em plus 0.5em minus 0.4em\relax Cham: Springer International Publishing,
  2016, pp. 154--169.

\bibitem{jointtrans}
H.~Zhang, V.~Sindagi, and V.~M. Patel, ``Joint transmission map estimation and
  dehazing using deep networks,'' \emph{IEEE Transactions on Circuits and
  Systems for Video Technology}, vol.~30, no.~7, pp. 1975--1986, 2020.

\bibitem{li2017allinone}
B.~Li, X.~Peng, Z.~Wang, J.~Xu, and D.~Feng, ``An all-in-one network for
  dehazing and beyond,'' \emph{arXiv preprint arXiv:1707.06543}, 2017.

\bibitem{zhu2015cap}
Q.~Zhu, J.~Mai, and L.~Shao, ``A fast single image haze removal algorithm using
  color attenuation prior,'' \emph{IEEE Transactions on Image Processing},
  vol.~24, no.~11, pp. 3522--3533, 2015.

\bibitem{berman2016non}
D.~Berman, S.~Avidan \emph{et~al.}, ``Non-local image dehazing,'' in
  \emph{Proceedings of the IEEE conference on computer vision and pattern
  recognition}, 2016, pp. 1674--1682.

\bibitem{li2015cod}
\BIBentryALTinterwordspacing
J.~Li, H.~Zhang, D.~Yuan, and M.~Sun, ``Single image dehazing using the change
  of detail prior,'' \emph{Neurocomputing}, vol. 156, pp. 1--11, 2015.
  [Online]. Available:
  \url{https://www.sciencedirect.com/science/article/pii/S0925231215000478}
\BIBentrySTDinterwordspacing

\bibitem{bui2017colorellipsoid}
T.~M. Bui and W.~Kim, ``Single image dehazing using color ellipsoid prior,''
  \emph{IEEE Transactions on Image Processing}, vol.~27, no.~2, pp. 999--1009,
  2018.

\bibitem{cai2016dehazenet}
B.~Cai, X.~Xu, K.~Jia, C.~Qing, and D.~Tao, ``Dehazenet: An end-to-end system
  for single image haze removal,'' \emph{IEEE Transactions on Image
  Processing}, vol.~25, no.~11, pp. 5187--5198, 2016.

\bibitem{zhang2018dcpdn}
H.~Zhang and V.~M. Patel, ``Densely connected pyramid dehazing network,'' in
  \emph{Proceedings of the IEEE conference on computer vision and pattern
  recognition}, 2018, pp. 3194--3203.

\bibitem{engin2018cycledehaze}
D.~Engin, A.~Genc, and H.~K. Ekenel, ``Cycle-dehaze: Enhanced cyclegan for
  single image dehazing,'' in \emph{2018 IEEE/CVF Conference on Computer Vision
  and Pattern Recognition Workshops (CVPRW)}, 2018, pp. 938--9388.

\bibitem{qin2020ffa}
X.~Qin, Z.~Wang, Y.~Bai, X.~Xie, and H.~Jia, ``Ffa-net: Feature fusion
  attention network for single image dehazing.'' in \emph{AAAI}, 2020, pp.
  11\,908--11\,915.

\bibitem{singh2020bppnet}
A.~Singh, A.~Bhave, and D.~K. Prasad, ``Single image dehazing for a variety of
  haze scenarios using back projected pyramid network,'' in \emph{European
  Conference on Computer Vision}.\hskip 1em plus 0.5em minus 0.4em\relax
  Springer, 2020, pp. 166--181.

\bibitem{pei2012nighttime}
S.-C. Pei and T.-Y. Lee, ``Nighttime haze removal using color transfer
  pre-processing and dark channel prior,'' in \emph{2012 19th IEEE
  International Conference on Image Processing}.\hskip 1em plus 0.5em minus
  0.4em\relax IEEE, 2012, pp. 957--960.

\bibitem{li2015nighttime}
Y.~Li, R.~T. Tan, and M.~S. Brown, ``Nighttime haze removal with glow and
  multiple light colors,'' in \emph{Proceedings of the IEEE international
  conference on computer vision}, 2015, pp. 226--234.

\bibitem{ancuti2016night}
C.~Ancuti, C.~O. Ancuti, C.~De~Vleeschouwer, and A.~C. Bovik, ``Night-time
  dehazing by fusion,'' in \emph{2016 IEEE International Conference on Image
  Processing (ICIP)}.\hskip 1em plus 0.5em minus 0.4em\relax IEEE, 2016, pp.
  2256--2260.

\bibitem{park2016nighttime}
D.~Park, D.~K. Han, and H.~Ko, ``Nighttime image dehazing with local
  atmospheric light and weighted entropy,'' in \emph{2016 IEEE International
  Conference on Image Processing (ICIP)}.\hskip 1em plus 0.5em minus
  0.4em\relax IEEE, 2016, pp. 2261--2265.

\bibitem{zhang2017fast}
J.~Zhang, Y.~Cao, S.~Fang, Y.~Kang, and C.~Wen~Chen, ``Fast haze removal for
  nighttime image using maximum reflectance prior,'' in \emph{Proceedings of
  the IEEE conference on computer vision and pattern recognition}, 2017, pp.
  7418--7426.

\bibitem{zhang2020nighttime}
J.~Zhang, Y.~Cao, Z.-J. Zha, and D.~Tao, ``Nighttime dehazing with a synthetic
  benchmark,'' in \emph{Proceedings of the 28th ACM International Conference on
  Multimedia}, 2020, pp. 2355--2363.

\bibitem{yu2019nighttime}
T.~Yu, K.~Song, P.~Miao, G.~Yang, H.~Yang, and C.~Chen, ``Nighttime single
  image dehazing via pixel-wise alpha blending,'' \emph{IEEE Access}, vol.~7,
  pp. 114\,619--114\,630, 2019.

\bibitem{lou2020integrating}
W.~Lou, Y.~Li, G.~Yang, C.~Chen, H.~Yang, and T.~Yu, ``Integrating haze density
  features for fast nighttime image dehazing,'' \emph{IEEE Access}, vol.~8, pp.
  113\,318--113\,330, 2020.

\bibitem{he2020side}
R.~He, X.~Guo, and Z.~Shi, ``Side—a unified framework for simultaneously
  dehazing and enhancement of nighttime hazy images,'' \emph{Sensors}, vol.~20,
  no.~18, p. 5300, 2020.

\bibitem{feng2020learning}
M.~Feng, T.~Yu, M.~Jing, and G.~Yang, ``Learning a convolutional autoencoder
  for nighttime image dehazing,'' \emph{Information}, vol.~11, no.~9, p. 424,
  2020.

\bibitem{kim1997bihist}
Y.-T. Kim, ``Contrast enhancement using brightness preserving bi-histogram
  equalization,'' \emph{IEEE Transactions on Consumer Electronics}, vol.~43,
  no.~1, pp. 1--8, 1997.

\bibitem{chen2003bihist}
S.-D. Chen and A.~Ramli, ``Minimum mean brightness error bi-histogram
  equalization in contrast enhancement,'' \emph{IEEE Transactions on Consumer
  Electronics}, vol.~49, no.~4, pp. 1310--1319, 2003.

\bibitem{ibrahim2007histeq}
H.~Ibrahim and N.~S. Pik~Kong, ``Brightness preserving dynamic histogram
  equalization for image contrast enhancement,'' \emph{IEEE Transactions on
  Consumer Electronics}, vol.~53, no.~4, pp. 1752--1758, 2007.

\bibitem{singh2015histeq}
\BIBentryALTinterwordspacing
K.~Singh, R.~Kapoor, and S.~K. Sinha, ``Enhancement of low exposure images via
  recursive histogram equalization algorithms,'' \emph{Optik}, vol. 126,
  no.~20, pp. 2619--2625, 2015. [Online]. Available:
  \url{https://www.sciencedirect.com/science/article/pii/S003040261500532X}
\BIBentrySTDinterwordspacing

\bibitem{land1977retinex}
E.~H. Land, ``The retinex theory of color vision,'' \emph{Scientific american},
  vol. 237, no.~6, pp. 108--129, 1977.

\bibitem{park2017llretinex}
S.~Park, S.~Yu, B.~Moon, S.~Ko, and J.~Paik, ``Low-light image enhancement
  using variational optimization-based retinex model,'' \emph{IEEE Transactions
  on Consumer Electronics}, vol.~63, no.~2, pp. 178--184, 2017.

\bibitem{li2018llretinex}
M.~Li, J.~Liu, W.~Yang, X.~Sun, and Z.~Guo, ``Structure-revealing low-light
  image enhancement via robust retinex model,'' \emph{IEEE Transactions on
  Image Processing}, vol.~27, no.~6, pp. 2828--2841, 2018.

\bibitem{zhang2019kindling}
\BIBentryALTinterwordspacing
Y.~Zhang, J.~Zhang, and X.~Guo, ``Kindling the darkness: A practical low-light
  image enhancer,'' in \emph{Proceedings of the 27th ACM International
  Conference on Multimedia}, ser. MM '19.\hskip 1em plus 0.5em minus
  0.4em\relax New York, NY, USA: ACM, 2019, pp. 1632--1640. [Online].
  Available: \url{http://doi.acm.org/10.1145/3343031.3350926}
\BIBentrySTDinterwordspacing

\bibitem{Chen2018DeepRetinex}
C.~Wei, W.~Wang, W.~Yang, and J.~Liu, ``Deep retinex decomposition for
  low-light enhancement,'' in \emph{British Machine Vision Conference}, 2018.

\bibitem{zhang2021kindplus}
\BIBentryALTinterwordspacing
Y.~Zhang, X.~Guo, J.~Ma, W.~Liu, and J.~Zhang, ``Beyond brightening low-light
  images,'' \emph{International Journal of Computer Vision}, vol. 129, no.~4,
  pp. 1013--1037, Apr 2021. [Online]. Available:
  \url{https://doi.org/10.1007/s11263-020-01407-x}
\BIBentrySTDinterwordspacing

\bibitem{guo2020zero}
C.~Guo, C.~Li, J.~Guo, C.~C. Loy, J.~Hou, S.~Kwong, and R.~Cong,
  ``Zero-reference deep curve estimation for low-light image enhancement,'' in
  \emph{Proceedings of the IEEE/CVF Conference on Computer Vision and Pattern
  Recognition}, 2020, pp. 1780--1789.

\bibitem{li2019benchmarking}
B.~Li, W.~Ren, D.~Fu, D.~Tao, D.~Feng, W.~Zeng, and Z.~Wang, ``Benchmarking
  single-image dehazing and beyond,'' \emph{IEEE Transactions on Image
  Processing}, vol.~28, no.~1, pp. 492--505, 2019.

\bibitem{ronneberger2015unet}
O.~Ronneberger, P.~Fischer, and T.~Brox, ``U-net: Convolutional networks for
  biomedical image segmentation,'' in \emph{International Conference on Medical
  image computing and computer-assisted intervention}.\hskip 1em plus 0.5em
  minus 0.4em\relax Springer, 2015, pp. 234--241.

\bibitem{guo2016lime}
X.~Guo, Y.~Li, and H.~Ling, ``Lime: Low-light image enhancement via
  illumination map estimation,'' \emph{IEEE Transactions on image processing},
  vol.~26, no.~2, pp. 982--993, 2016.

\bibitem{chan2011augmented}
S.~H. Chan, R.~Khoshabeh, K.~B. Gibson, P.~E. Gill, and T.~Q. Nguyen, ``An
  augmented lagrangian method for total variation video restoration,''
  \emph{IEEE Transactions on Image Processing}, vol.~20, no.~11, pp.
  3097--3111, 2011.

\bibitem{huang2017densenet}
G.~Huang, Z.~Liu, L.~Van Der~Maaten, and K.~Q. Weinberger, ``Densely connected
  convolutional networks,'' in \emph{Proceedings of the IEEE conference on
  computer vision and pattern recognition}, 2017, pp. 4700--4708.

\bibitem{simonyan2014vgg}
K.~Simonyan and A.~Zisserman, ``Very deep convolutional networks for
  large-scale image recognition,'' \emph{arXiv preprint arXiv:1409.1556}, 2014.

\bibitem{Johnson2016Perceptual}
J.~Johnson, A.~Alahi, and L.~Fei-Fei, ``Perceptual losses for real-time style
  transfer and super-resolution,'' in \emph{European conference on computer
  vision}.\hskip 1em plus 0.5em minus 0.4em\relax Springer, 2016, pp. 694--711.

\bibitem{santra2016dnu}
S.~Santra and B.~Chanda, ``Day/night unconstrained image dehazing,'' in
  \emph{2016 23rd International Conference on Pattern Recognition (ICPR)},
  2016, pp. 1406--1411.

\end{thebibliography}
